\documentclass[10pt,twocolumn,letterpaper]{article}

\usepackage{iccv}
\usepackage{times}
\usepackage{epsfig}
\usepackage{graphicx}
\usepackage{amsmath}
\usepackage{amssymb}

\usepackage{graphicx}
\usepackage{amsmath}
\usepackage{amssymb}
\usepackage{booktabs}
\usepackage{multirow}
\usepackage[accsupp]{axessibility}

\newcommand{\new}[1]{#1}

\usepackage[pagebackref=false,breaklinks=true,letterpaper=true,colorlinks,bookmarks=false]{hyperref}

\iccvfinalcopy %
\usepackage[capitalize]{cleveref}
\crefname{section}{Sec.}{Secs.}
\Crefname{section}{Section}{Sections}
\Crefname{table}{Table}{Tables}
\crefname{table}{Tab.}{Tabs.}

\ificcvfinal\fi

\begin{document}

\title{Multi-Modal Neural Radiance Field for Monocular Dense SLAM \\ with a Light-Weight ToF Sensor}

\author{
  Xinyang Liu$^{1}$,
  Yijin Li$^{1}$,
  Yanbin Teng$^{1}$,
  Hujun Bao$^{1}$,
  Guofeng Zhang$^{1}$,
  Yinda Zhang$^{2}$,
  Zhaopeng Cui$^{1}$\thanks{Corresponding author.}\and
  \textnormal{$^1$State Key Lab of CAD\&CG, Zhejiang University \quad $^2$Google} 
}

\maketitle
\ificcvfinal\thispagestyle{empty}\fi

\begin{abstract}
Light-weight time-of-flight (ToF) depth sensors are compact and 
cost-efficient, and thus widely used on mobile devices 
for tasks such as autofocus and obstacle detection.
However, due to the sparse and noisy depth measurements, these sensors have rarely been considered for dense geometry reconstruction.
In this work, we present the first dense SLAM system with a monocular camera and a light-weight ToF sensor.
Specifically, we propose a multi-modal implicit scene representation that supports 
rendering both the signals from the RGB camera and light-weight ToF sensor which drives the optimization by comparing with the raw sensor inputs.
Moreover, in order to guarantee successful pose tracking and reconstruction, we exploit a predicted depth as an intermediate supervision and develop a coarse-to-fine optimization strategy for efficient learning of the implicit representation.
At last, the temporal information is explicitly exploited to deal with the noisy signals from light-weight ToF sensors to improve the accuracy and robustness of the system. 
Experiments demonstrate that our system well exploits the signals of light-weight ToF sensors and achieves competitive results both on camera tracking and dense scene reconstruction. Project page: \url{https://zju3dv.github.io/tof_slam/}.

\end{abstract}

\section{Introduction}
Dense simultaneous localization and mapping (dense SLAM)~\cite{whelanElasticFusionRealtimeDense2016,daiBundleFusionRealtimeGlobally2017,zilong_slam,consistent_depth} has extensive applications in augmented reality~\cite{ptam,rk_slam}, indoor robotics, etc.
It usually relies on high-precision and high-resolution depth sensors, such as time-of-flight (ToF) sensors or structured light sensors.
\new{
Due to the size, weight, and price issues, these depth sensors are only used in a few high-end mobile devices until recent years. 
In contrast, light-weight ToF sensors, which are cost-effective, compact, and energy-efficient, were integrated into hundreds of smartphone models\footnote{\label{footnote:st}\url{https://www.st.com/content/st_com/en/about/media-center/press-item.html/t4210.html}}.
As a result, it would be valuable if we could fully utilize these light-weight sensors for dense SLAM, which further facilitates other applications like AR/VR and micro-Robot.
}

\begin{figure}[!t]
\begin{center}
\includegraphics[width=1.0\linewidth]{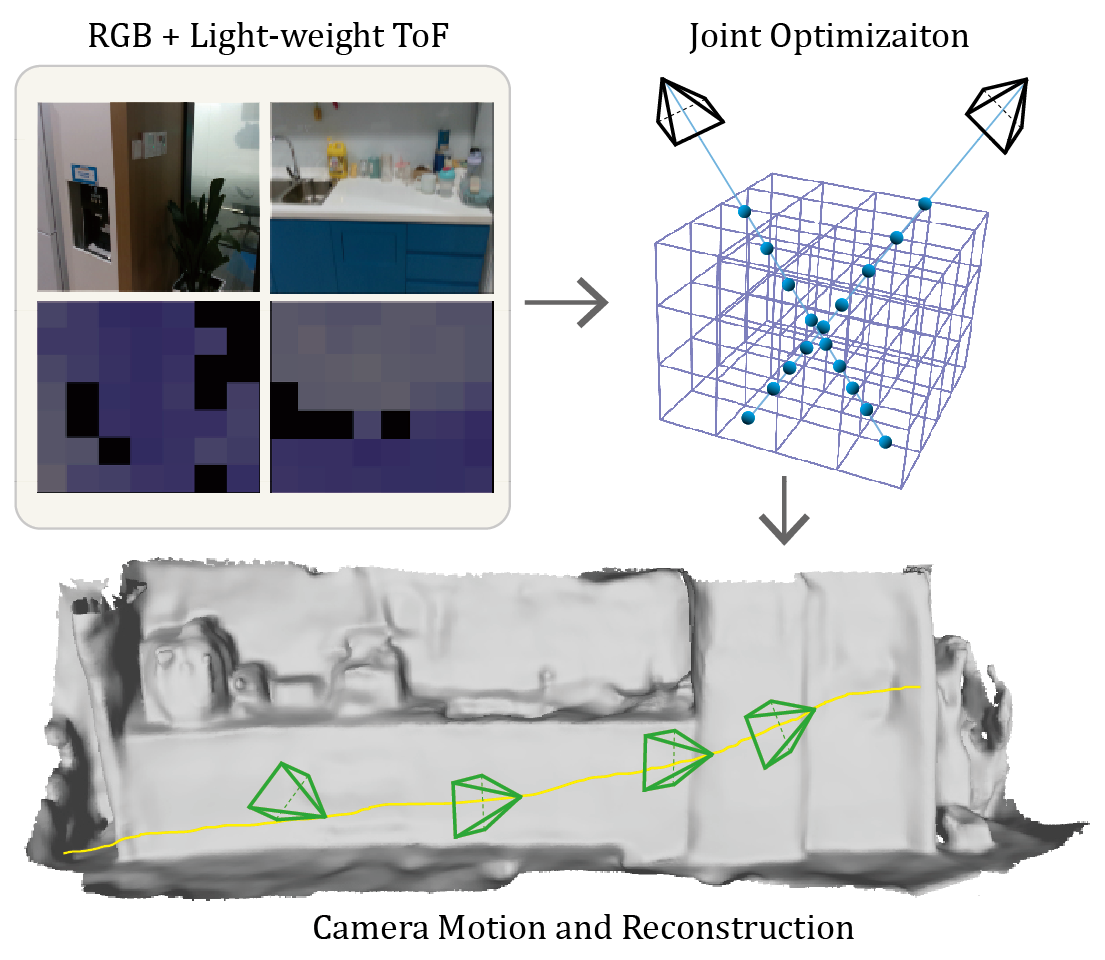}
\end{center}
\caption{\new{\textbf{Monocular Dense SLAM with Our Multi-Modal Implicit Representation.}}
We present a novel SLAM system based on implicit scene representation. The system does not require high-precision and high-resolution depth sensors and only takes RGB images and the signals of light-weight ToF sensors as input.
}
\label{fig:teaser}
\end{figure}

\begin{figure}[!t]
\begin{center}
    \includegraphics[width=0.9\linewidth]{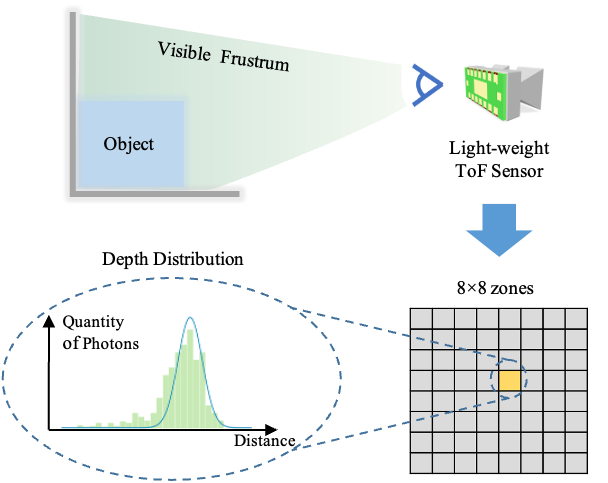}
\end{center}
\caption{\textbf{L5 Sensing Principle.} Instead of pixel-wise depth measurement, L5 measures the depth distribution of a large zone and returns its depth mean and variance. Specifically, L5 measures $8\times 8$ zones in total. The figure is adapted from~\cite{li2022deltar}.}
\label{fig:principle}
\end{figure}

Unfortunately, limited by the compact electronic design, the light-weight ToF sensor can only provide coarse measurement in the form of depth distribution in an extremely low resolution as illustrated in Fig.~\ref{fig:principle}. 
Existing RGB-D dense SLAM systems~\cite{whelanElasticFusionRealtimeDense2016,zhuNICESLAMNeuralImplicita,sucar2021imap} are designed for accurate and pixel-wise depth inputs, thus cannot work with the light-weight ToF signals directly. They will also fail if we simply consider the light-weight ToF signals as a low-resolution depth (\ie, mean depth values in each zone).

In this paper, we aim to design a novel learning-based dense SLAM system that provides accurate pose tracking and dense reconstruction taking the RGB sequences from a color camera and the sparse signals of light-weight ToF as input (Fig.~\ref{fig:teaser}). 
However, it is non-trivial to design such a system. At first, 
motivated by the recent achievements in the field of neural rendering
~\cite{mildenhallNerfRepresentingScenes2020, barronMipNeRFMultiscaleRepresentation2021,neural_rendering_in_a_room,neumesh,obj_nerf} and grid-based feature encoding~\cite{zhuNICESLAMNeuralImplicita,wangGOSurfNeuralFeature2022,nsvf}, we propose to design our system based on a novel implicit representation from which we can render both the RGB image and original ToF signal. In this way, we can define the losses directly against the multi-domain input, which can be utilized to optimize camera poses and 3D scenes.
However, we find that this cannot guarantee plausible tracking and reconstruction results because of low-quality raw depth signals from the sensor. Inspired by recent works~\cite{sucar2021imap,zhuNICESLAMNeuralImplicita} that high-resolution accurate depth maps play an important role in the implicit SLAM systems, we further exploit the depth estimation model~\cite{li2022deltar} for the light-weight ToF sensor to predict an intermediate high-resolution depth as additional supervision.
Based on all the above insights, we develop a multi-modal implicit scene representation with multi-level feature grids, which is able to generate both the zone-level signals of light-weight ToF sensors and pixel-wise RGB/depth images via the differentiable rendering for the camera tracking and reconstruction. 
For efficient network convergence, we also design a coarse-to-fine optimization process for this novel implicit scene representation, \ie, firstly using zone-level ToF signals to optimize the scene at the coarse level, then adding pixel-wise RGB/depth supervisions to recover geometry details.

Moreover, although the predicted per-pixel depth is generally smooth and provides reasonable supervision as intermediate signals, it may also produce severe artifacts when there is a large portion of missing L5 signals since it is hard for the network to handle such cases due to the inherent depth ambiguity in the missing regions. As a result, we further develop a temporal filtering technique to enhance depth prediction.
Specifically, when a new signal is captured, we render a zone-level light-weight ToF signal from our multi-modal scene representation with an initialized pose and fuse it with that new observation signal, which serves as the input of the depth prediction network. 
Such an explicit filtering technique improves the depth estimation performance significantly, particularly in extreme cases where the raw L5 signals are very noisy or contain large amounts of missing data, and therefore further benefits the whole SLAM system.

Our contributions can be summarized as follows.
At first, to our best knowledge, we present the first dense SLAM system by only taking the monocular images and the signals from a light-weight ToF sensor as input.
Moreover, we propose a multi-modal implicit scene representation which supports rendering both the zone-level signals of light-weight ToF sensors and pixel-wise RGB/depth images. By minimizing the re-rendering loss of these signals in a coarse-to-fine strategy, we can recover the camera pose and the scene geometry via differentiable neural rendering.
Furthermore, we propose a temporal filtering technique to enhance the signals of light-weight ToF sensors and corresponding depth prediction which significantly improve the proposed SLAM system in extreme cases.
Experiments on the real datasets demonstrate that the proposed system well exploits the signals of light-weight ToF sensors and achieves competitive results both on camera tracking and dense scene reconstruction compared to existing methods.

\section{Related Work}
\noindent\textbf{Visual SLAM.}
Sparse visual SLAM systems~\cite{ptam,mur2017orb,ice_ba} focus on solving accurate camera poses based on the tracking of sparse keypoint~\cite{rublee2011orb,pats}. These types of methods usually struggle in texture-less environments and cannot provide a complete reconstruction result of the scene.
In comparison, dense visual SLAM systems~\cite{izadi2011kinectfusion,whelanElasticFusionRealtimeDense2016,kerl2013dense,newcombe2011dtam} perform much more robust but usually require depth images as input.
More recently, many methods emerged to estimate the dense depth map and camera pose simultaneously from RGB sequences only using deep neural networks~\cite{bloesch2018codeslam,czarnowski2020deepfactors,teed2019deepv2d,teed2021droid} and reconstruct the scene through fusing the estimated depth maps.
Unlike these previous methods optimizing depth maps per frame, we follow iMAP~\cite{sucar2021imap} and NICE-SLAM~\cite{zhuNICESLAMNeuralImplicita} and use an implicit scene representation. However, unlike iMAP and NICE-SLAM, we do not require the depth camera and take the light-weight ToF signals as input instead.

\begin{figure*}[!t]
\begin{center}
    \includegraphics[width=1.0\linewidth]{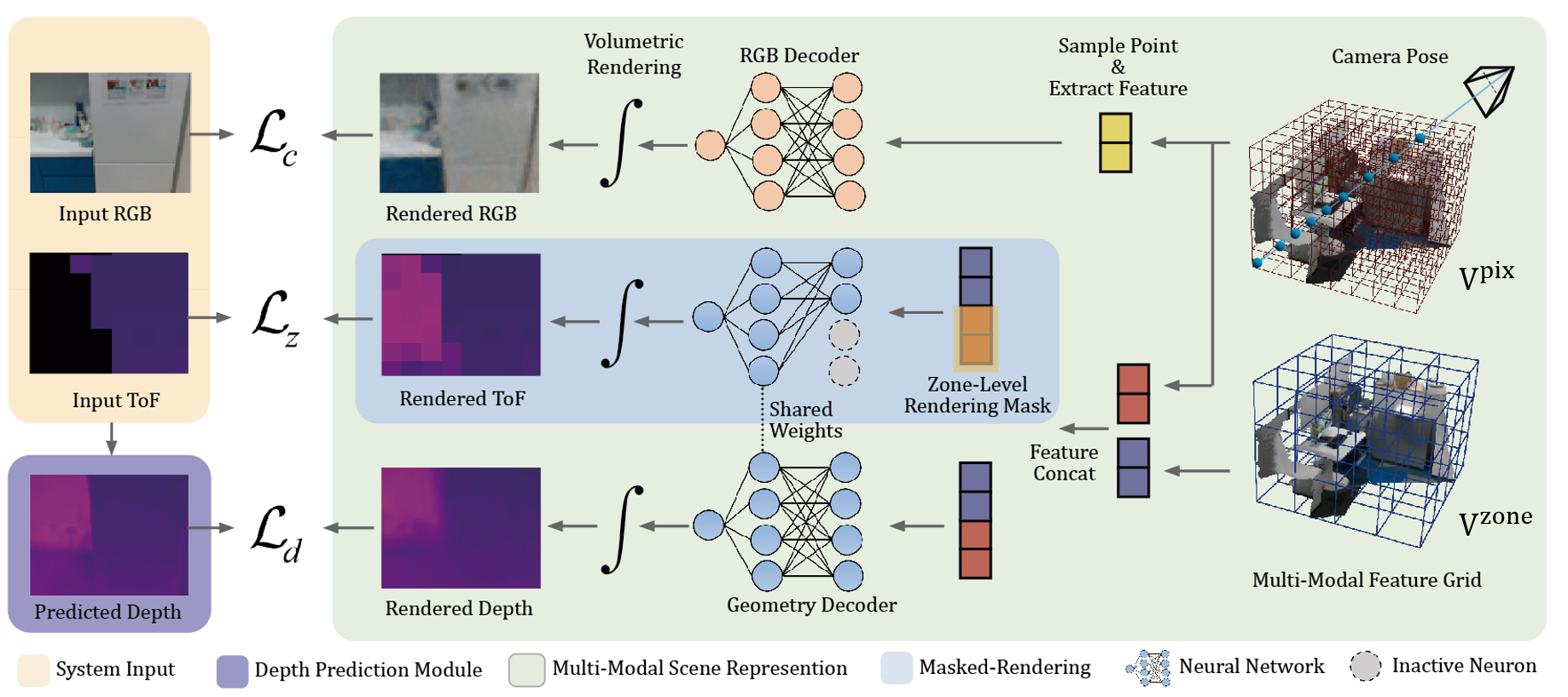}
\end{center}
\caption{\textbf{System Overview.} Our method uses a monocular camera and a light-weight ToF sensor as input and recovers both camera motion and the scene structures. Through differentiable rendering, our method can render multi-modal signals, including color images, depth images and zone-level L5 signals. Both the scene structure and the camera poses are optimized by minimizing the re-rendering loss.}
\label{fig:pipeline}
\end{figure*}

\noindent\textbf{Neural Implicit Representation.}
Neural implicit representations are widely used in various kinds of tasks, including novel view synthesis~\cite{mildenhallNerfRepresentingScenes2020, zhang2020nerf++,barronMipNeRFMultiscaleRepresentation2021,martin2021nerf} and scene reconstruction~\cite{azinovicNeuralRGBDSurface,wangGOSurfNeuralFeature2022,ortizISDFRealTimeNeural2022}. While showing promising results, these methods require precise camera poses as input which greatly limits their application scenarios. Some recent works~\cite{linBARFBundleAdjustingNeurala,yen-chenINeRFInvertingNeural2021} propose to optimize camera pose with the scene representation 
simultaneously.
Lately, iMAP~\cite{sucar2021imap} and NICE-SLAM~\cite{zhuNICESLAMNeuralImplicita} bring neural implicit representations into visual SLAM systems. 
Nevertheless, requirements of RGB-D sequences as input significantly constrain their applications, especially on mobile devices with tight power budgets. In contrast, we propose a visual SLAM method using only an RGB camera and a light-weight ToF sensor, which commonly co-exists on mobile devices.

\noindent\textbf{Neural Grid Optimization.}
Typically, a single multi-layer perceptron (MLP) is employed to represent the entire scene in earlier neural implicit methods~\cite{mildenhallNerfRepresentingScenes2020, barronMipNeRFMultiscaleRepresentation2021,sucar2021imap}.
However, these methods perform poorly when applied to large scenes and need an extremely long time for training. To address this issue, recent works~\cite{takikawa2021neural,instant_ngp,chen2022tensorf} try to optimize neural features on an explicit volumetric grid. 
These methods, however, are only capable of pixel-level rendering and do not accommodate L5 signals as it has extremely low resolution and measures the depth distribution of a large area.
To this end, we introduce a multi-modal feature grid representation that enables rendering multi-modal information at different resolutions and can deal with zone-level depth distribution data measured by light-weight ToF sensors.

\section{Method}

In this section, we provide an overview of the sensing principle of light-weight ToF sensors (\eg, L5) in Sec.~\ref{sec:l5-principle}.
Based on the characteristics of L5's signals, we design the first dense SLAM system with a monocular camera and a light-weight ToF sensor as shown in Fig.~\ref{fig:pipeline}.
Specifically, we first propose a multi-modal implicit scene representation which enables rendering L5 signals together with the common RGB image and depth maps (Sec.~\ref{sec:multi-modal}). To guarantee successful tracking and mapping, we exploit a depth prediction model~\cite{li2022deltar} to predict intermediate per-pixel depth maps as additional supervision.  By minimizing the difference between the rendered signals and input/predicted ones, we can simultaneously optimize the camera pose and scene structure in a coarse-to-fine way.
Furthermore, since the depth prediction may contain severe artifacts when there is a large portion of missing L5 signals, we propose to refine the L5 signals with temporal filtering techniques to enhance the depth prediction module (Sec.~\ref{sec:depth predicton}).
Finally, we describe the system implementation detail in Sec.~\ref{sec:slam_system}.

\subsection{Preliminaries: L5 Sensing Principle}
\label{sec:l5-principle}
Light-weight ToF sensors are designed to be low-cost, small, and low-energy and have been massively deployed on mobile devices. Compared with conventional ToF sensors, which provide high-resolution depth measurement 
and measure the per-pixel distance to the scene, light-weight ToF sensors usually have extremely low resolution (e.g., $8\times 8$ zones) and measure the depth distribution for each zone.
Here we take ST VL53L5CX~\cite{l5_page} (denoted as L5), a representative product of light-weight ToF sensors, as an example to declare the sensing principle of these sensors.
As shown in Fig.~\ref{fig:principle}, L5 measures the depth distribution by counting 
the received photon number within specific time intervals.
The result is then fitted with a Gaussian distribution, and L5 only transmits the mean and variance 
to decrease both the energy consumption and broadband load.
None of the previous studies have explored the usage of L5 for downstream applications like SLAM due to its low resolution and high uncertainty.

\subsection{Multi-Modal Implicit Scene Representation}
\label{sec:multi-modal}

The combination of neural rendering and grid-based feature encoding has found broad utilization in applications like SLAM and surface reconstruction.~\cite{zhuNICESLAMNeuralImplicita,wangGOSurfNeuralFeature2022,yuMonoSDFExploringMonocular2022}.
By minimizing the loss between the rendered image/depth and the input image/depth, they achieve accurate 6-DoF camera pose tracking and the reconstruction of scene geometry.
In this paper, we propose a multi-modal implicit scene representation, which supports the rendering of zone-level L5 signals apart from the common RGB and depth images.
Specifically, we encode the geometry and color separately and propose the masked rendering technique in the geometry encoding for the rendering of both zone-level L5 signals and pixel-level depth images.

\noindent\textbf{Geometry Encoding with Masked Rendering.}
The idea of the proposed masked rendering is inspired by the integrated positional encoding(IPE) theory proposed in Mip-NeRF~\cite{barronMipNeRFMultiscaleRepresentation2021}, but we promote it to the grid-based scene representation. The core idea of IPE is that the input features are passed through a low pass filter, \ie, if the frequency of a particular feature has a larger period than the ray, then the feature is unaffected; otherwise, the feature is scaled down towards zero. The original method represents the scene using a single MLP and achieves the low pass filtering by calculating an integral of the positional encoding as input. In our grid-based case, we concatenate features from different-level feature grids and use a rendering mask to mask out features extracted from overly high spatial frequency grids based on the current rendering scale. 

To be more specific, we encode the scene geometry into a multi-level feature grid containing four layers ${{\cal V}_\theta } = \{ {V^0},{V^1},{V^2},{V^3}\} $. The zone-level feature grid ${V^{\text{zone}}}$ contains the coarser feature grids$\{ {V^0},{V^1}\}$  while the pixel-level feature grid  ${V^{\text{pix}}}$ contains the finer feature grids$\{ {V^2},{V^3}\}$. For a given point ${\bf{x}} \in {\mathbb{R}^3}$, its whole geometry feature $\cal{F}$ is extracted by tri-linearly interpolating features at each grid level and concatenating these features together. The feature can be decoded into SDF values in both pixel-level $\phi_{pix}\left(\mathbf{x}\right)$ and zone-level $\phi_{zone}\left(\mathbf{x}\right)$  via the same geometry decoder $ {f_\omega }\left( \cdot \right)\ $ switched by the rendering mask.
We use all the features for rendering pixel-level results $\phi _{pix}$, and mask out the features extracted from the finer grids for zone-level rendering results $\phi _{zone}$:
\begin{equation}
\begin{split}
    {\phi _{pix}}({\bf{x}}) &= {f_\omega }\left( {[{V^{zone}}({\bf{x}}),{V^{pix}}({\bf{x}})]} \right),\\
    {\phi _{zone}}({\bf{x}}) &= {f_\omega }\left( {[{V^{zone}}({\bf{x}}),{\bf{0}}]} \right).
\end{split}
\end{equation}
This mask operation also makes the corresponding neurons in the geometry decoder inactive.
\new{
The SDF values from both branches are used in the tracking and mapping process corresponding to the pixel-level and zone-level supervision, and only the pixel-level SDF values are used for the final mesh extraction.
}

\noindent\textbf{Color Encoding.}
For color information, we encode it only at the finest level using a separate set of feature grid $\cal W_\beta$ and decoder $g_\gamma \left( \cdot \right)$ as~\cite{zhuNICESLAMNeuralImplicita, wangGOSurfNeuralFeature2022}. When decoding color, we additionally use the ray direction $\mathbf{r}$, so the color value for a 3D point is given by:
\begin{equation}
{\bf{c}} = {g_\gamma }\left( {{{\cal W}_\beta }({\bf{x}}),{\bf{r}}} \right).
\end{equation}

\noindent\textbf{Rendering of L5 Signals, Color and Depth Images.}
We render color and depth values with the volumetric rendering technique~\cite{wangNeuSLearningNeural2021}.
Specifically, to render a color pixel, we sample $N$ points along the corresponding emitted ray, denoted as ${{\bf{x}}_i} = {\bf{o}} + {d_i}{\bf{r}},i \in {\rm{\{ 1,2,}}...{\rm{,N\} }}$ where $\bf{o}$ is the camera center, $\bf{r}$ is the direction of this ray and $d_i$ is the distance of the sampled point $\bf{x}_i$ along the ray.
We then accumulate the color value along the ray through:
\begin{equation}
\label{eq:color-render}
\hat {\bf{c}} = \sum\limits_{i = 1}^N {T_i}{\alpha _i} {{\bf{c}}_i},
\end{equation}
where ${T_i} = \prod\limits_{j = 1}^{i - 1} {\left( {1 - {\alpha _j}} \right)}$ is the accumulated transmittance,
and $\alpha_i$ is the opacity value converted from the SDF prediction $\phi_{pix}$. The conversion follows the original definition in NeuS~\cite{wangNeuSLearningNeural2021}.

The process of rendering the L5 signals is similar to rendering the color.
The differences are that for rendering the mean depth value $\hat d_z$ of an L5 zone, we emit the ray from the center of that zone and accumulate the distance along the ray through:
\begin{equation}
\label{eq:depth-render}
\hat d_z = \sum\limits_{i = 1}^N {T_i}{\alpha _i} {d_i}.
\end{equation}
\new{
The zone-level opacity $\alpha _i$ is derived in the same manner as color rendering but using $\phi_{zone}$ instead of $\phi_{pix}$.
}
Intuitively, we can optimize the camera pose and scene structure by only supervising the rendered color images and L5 signals. 
However, in our experiment (Sec.~\ref{sec:ablation}), we show that the results are far from satisfactory.
As a result, we also render the pixel-wise depth maps using Eq.~\ref{eq:depth-render} with $\phi_{pix}$ and supervise it with the depth prediction from~\cite{li2022deltar}.

\subsection{Temporal Filtering of L5 Signals}
\label{sec:depth predicton}

As mentioned before, we use DELTAR~\cite{li2022deltar} to predict a pixel-wise depth map as additional supervision. DELTAR is a pre-trained neural network that takes L5 signals and RGB images as input and predicts corresponding depth maps.
We observe that when there are a large number of missing or noisy L5 signals, the depth map predicted by DELTAR may contain severe artifacts due to the inherent depth ambiguities of missing or noisy regions, thus further contaminating the learning of implicit features and degrading the SLAM system's performance. 

\begin{figure}[!t]
\begin{center}
\includegraphics[width=1.0\linewidth]{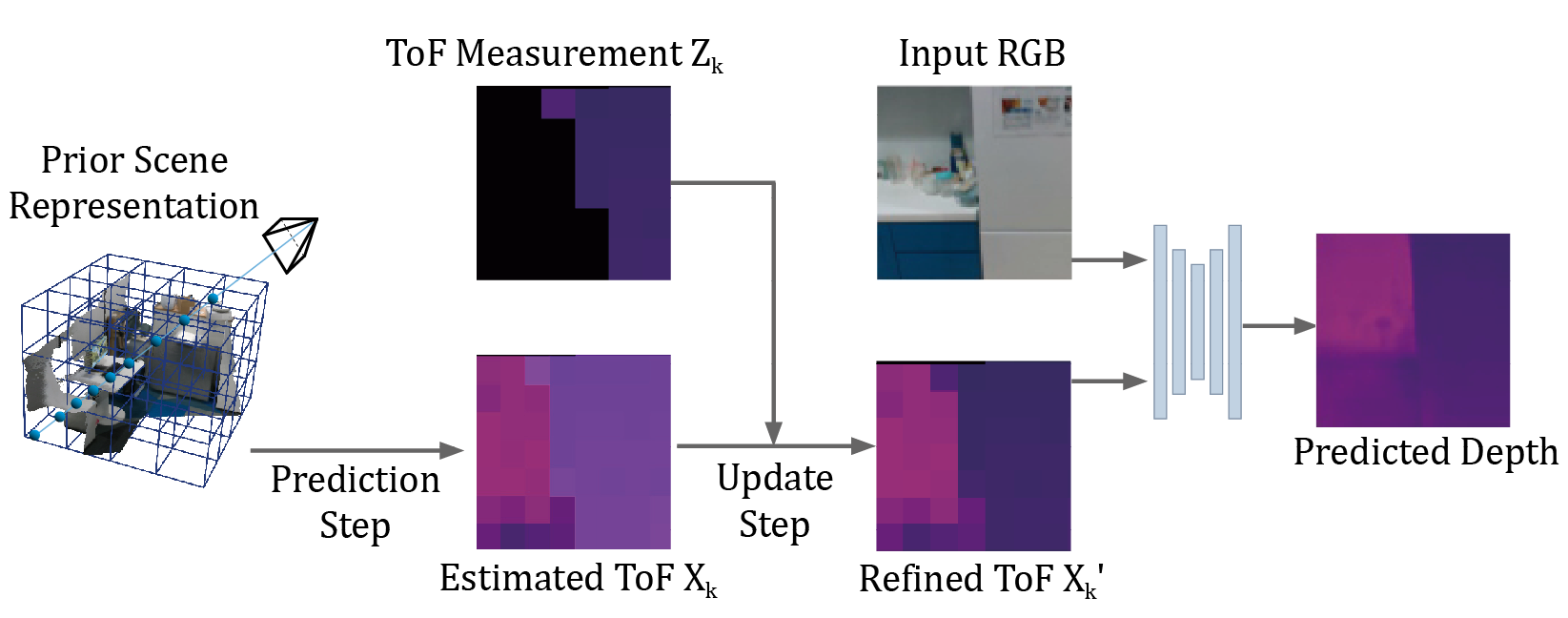}
\end{center}
\caption{\textbf{Temporal Filtering of L5 Signals.} By fusing the latest ToF measurement with a rendered one, we can significantly improve the signal quality and the corresponding predicted depth.}
\label{fig:temporal}
\end{figure}

This motivates us to develop an explicit temporal filtering technique (Fig.~\ref{fig:temporal}) to enhance the L5 signals before feeding into DELTAR.
Since the past observations are stored implicitly in our scene representation, we can leverage this information to refine the current observation.
Specifically, the proposed filtering algorithm contains two steps: the prediction step and the update step.
In the prediction step, we predict the per-zone ToF measurement $X_k=\{\mu _1,\sigma _1 \}$ in the timestamp $k$ using neural rendering (Eq.~\ref{eq:depth-render}) with an initialized pose.
Then we update $X_k$ to $X'_k$ with the current L5 measurement $Z_k=\{\mu _2,\sigma _2 \}$:
\begin{equation}
    {\mu _{{X_k}'}} = \frac{{{\mu _1}\sigma _2^2 + {\mu _2}\sigma _1^2}}{{\sigma _1^2 + \sigma _2^2}},\quad\sigma _{{X_k}'}^2 = \frac{{\sigma _1^2\sigma _2^2}}{{\sigma _1^2 + \sigma _2^2}}.
\end{equation}
As for the zones with no valid raw L5 signals, we simply use the predicted L5 signals. 
Finally, the enhanced signals $X'_k$ are fed into the depth prediction network with corresponding RGB image~\cite{li2022deltar} to obtain a high-resolution depth estimation.

\subsection{SLAM System Implementation}
\label{sec:slam_system}
\noindent\textbf{Initialization.}
Since we do not have reliable depth maps as input, we need to perform initialization in order to build a local map for our system to bootstrap. 
The first $N_i$ frames are added to the initialization process with a fixed interval $I_{skip}$ sequentially. 
The pose of each newly added frame is initialized by the previous frame. 
During this process, the feature grids, decoders and camera poses are jointly optimized using the same loss function as in the mapping process. 
Finally, when all the frames are added, we optimize the whole frame set for $N_e$ iterations.

\begin{figure*}[!t]
\begin{center}
\includegraphics[width=.95\linewidth]{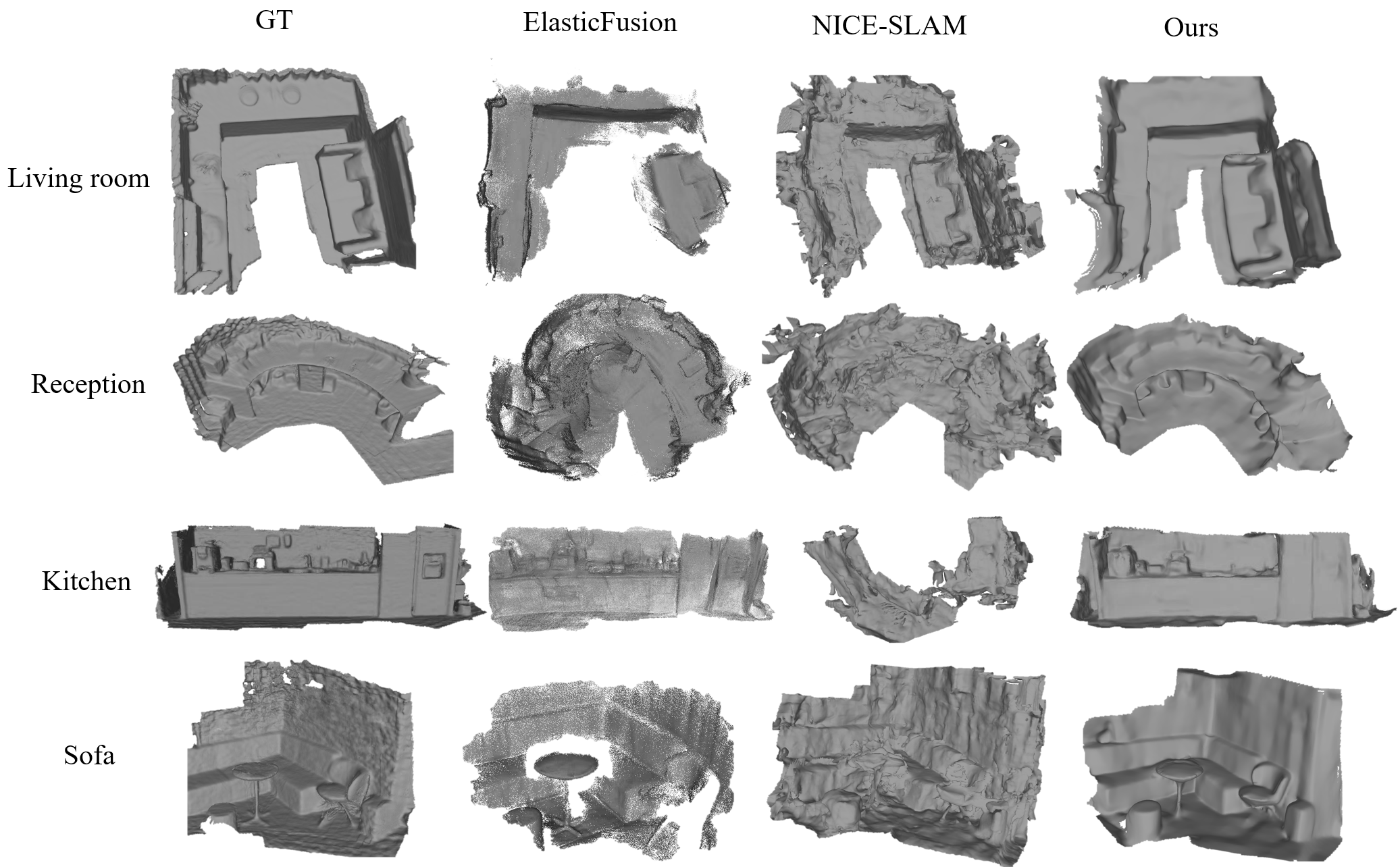}
\end{center}
\caption{\textbf{Qualitative Reconstruction Results.} Compared to other methods, our approach produces high-quality reconstructions with smooth surfaces and fewer artifacts.}
\label{fig:mesh}
\end{figure*}

\noindent\textbf{Coarse-to-Fine Mapping.}
\label{sec:mapping}
To optimize the scene representation mentioned in Sec.~\ref{sec:multi-modal}, we uniformly sample total $ M $ pixels and $ Z $ zones and perform optimization to minimize the photometric loss, geometric loss and SDF regularization. The zones and pixels are sampled from the current optimization window, which consists of two types of frames: neighbor frames (the nearest $N_n$ frames with an interval of 4) and global frames (chosen from all the past co-visible mapping frames).
\new{The photometric loss is the average color difference of all the sampled pixels:}
\begin{equation}
    {{\cal L}_{c}} = \frac{1}{M}\sum\limits_{m = 1}^M |I[u_m, v_m]-\hat{\mathbf{c}_m}|,
\end{equation}
where $u_m, v_m$ represents for the image coordinate of pixel $m$.
The geometric loss is calculated on both pixel-level and zone-level. For zone-level loss, only zones with valid L5 signals $Z_d$ are considered. We also add SDF supervision on pixel-level as in~\cite{wangGOSurfNeuralFeature2022,ortizISDFRealTimeNeural2022} to add robustness to the optimization. The whole geometric loss is defined as:
\begin{equation}
\begin{split}
{{\cal L}_g} &= {{\cal L}_d} + {{\cal L}_z} + {\lambda _{sdf}}{{\cal L}_{sdf}}, \\
{{\cal L}_d} &= \frac{1}{M}\sum\limits_{m = 1}^M {|\widetilde D[{u_m},{v_m}] - {{\hat d}_m}|}, \\
{{\cal L}_z} &= \frac{1}{{\left| {{Z_d}} \right|}}\sum\limits_{z \in {Z_d}} {|{{\bar D}_z} - {{\hat d}_z}|},
\end{split}
\end{equation}
where $ { \overline {{D}} } \in {\mathbb{R}^{8 \times 8}}$ represents the mean depth value measured by L5 and $\widetilde D$ represents the pixel-wise depth prediction.
Since we do not have high-quality depth as input and there are many texture-less areas in typical indoor scenes, directly optimizing the feature volume and camera poses is an ill-posed problem. Therefore, we employ two regularizations on the SDF values during pixel-level rendering as~\cite{wangGOSurfNeuralFeature2022}, namely the eikonal term $ {\cal L}_{eik} $ and the smoothness term $ {\cal L}_{s} $:

\begin{equation}
    {{\cal L}_{r}} = {\lambda _{eik}}{{\cal L}_{eik}} + {\lambda _{s}}{{\cal L}_{s}},
\end{equation}
\new{where the eikonal regularization is employed to ensure the network produces valid SDF values~\cite{wangNeuSLearningNeural2021, wangGOSurfNeuralFeature2022}, the smoothness regularization is used to encourage neighbor points to have similar normal directions, and $\lambda _{eik}$ and $\lambda _{s}$ are weights to balance the two regularization terms.}

During the mapping process, we perform a coarse-to-fine optimization process, bringing better convergence since pixel-wise optimization can rely on the already initialized coarse grids.
We first optimize the scene at the coarse level using zone-level ToF signals. Then, pixel-wise RGB/depth supervisions are added to jointly optimize the decoders, feature grids and camera poses by minimizing the mentioned losses for $N_m$ iterations with a local BA to cover geometry details. 
\begin{equation}
    \label{eq:mapping}
    {\min _{\theta ,\omega ,\beta ,\gamma ,\left\{ {{{\bf{R}}_i},{{\bf{t}}_i}} \right\}}}{\lambda _{c}}{{\cal L}_{c}} + {\lambda _{g}}{{\cal L}_{g}} + {\lambda _{r}}{{\cal L}_{r}}.
\end{equation}
We also conduct a global BA every $N_g$ frame by adding all the past mapping frames to the optimization window and jointly optimize them to globally optimize the camera poses and scene geometry.
We provide the details of losses in the supplementary material.

\noindent\textbf{Tracking.}
During tracking, the feature grid and the decoders remain fixed, 
while only the 6-DoF pose of the current frame is optimized.
Similar to previous methods~\cite{zhuNICESLAMNeuralImplicita,sucar2021imap}, 
$M_t$ pixels are sampled in the current frame to calculate both the pixel-level photometric loss and geometric loss as previously described.
The camera pose is then updated iteratively via back-propagation. Tracking with only pixel-level supervision is susceptible to a suboptimal solution, so we leverage our multi-modal scene representation and apply zone-level supervision in the early stage. Specifically, we additionally sample $ Z $ zones in the current frame and use the L5 signal to minimize zone depth loss as well, which leads to a more robust tracking result.

\begin{table*}[!t]
  \centering
  \small
    \begin{tabular}{llcccccccc}
    \toprule
    Scene Name                     &         & Kitchen        & Sofa           & Office         & Reception      & Living room    & Office2        & Sofa2          & Avg.           \\ \hline
    
    \multirow{3}{*}{KinectFusion\cite{izadi2011kinectfusion}} 
                                   & Acc.↓   & -    & 0.190          & 0.211          & 0.261          & -       & 0.267 & 0.135          & 0.213          \\
                                   & Comp.↓  & -  & 0.048          & 0.046          & 0.064            & -       & \textbf{0.078} &0.064         & \textbf{0.060}         \\
                                   & F-score & -    & 0.278         & 0.288      & 0.285          & -            & 0.274  & 0.381          & 0.301          \\\hline
                                   
    \multirow{3}{*}{ElasticFusion\cite{whelanElasticFusionRealtimeDense2016}} 
                                   & Acc.↓   & \underline{0.092}    & 0.135          & \underline{0.084}          & 0.297          & 0.151          & \textbf{0.096} & \underline{0.122}          & 0.140          \\
                                   & Comp.↓  & \textbf{0.065} & 0.048          & 0.082          & 0.305          & 0.216          & 0.147          & 0.047          & 0.130          \\
                                   & F-score & \underline{0.553}    & 0.420          & \underline{0.529}    & 0.274          & 0.382          & 0.416          & 0.481          & 0.436          \\\hline
    \multirow{3}{*}{BundleFusion\cite{daiBundleFusionRealtimeGlobally2017}}  
                                   & Acc.↓   & 0.170          & \underline{0.100}    & 0.103    & \underline{0.122}    & -              & 0.121          & 0.123    & \underline{0.123}    \\ 
                                   & Comp.↓  & 0.088          & \textbf{0.030} & \textbf{0.038} & \textbf{0.057} & -              & 0.214          & \underline{0.034}    & 0.077    \\
                                   & F-score & 0.373          & \underline{0.571}    & 0.474          & \underline{0.470}    & -              & \underline{0.442}    & \underline{0.527}    & \underline{0.476}    \\ \hline
    \multirow{3}{*}{iMAP\cite{sucar2021imap}}          
                                   & Acc.↓   & -              & 0.135          & 0.229          & 0.365          & 0.225          & 0.233          & 0.139         & 0.221          \\
                                   & Comp.↓  & -              & 0.054          & 0.103          & 0.245          & 0.291          &0.139          & 0.069          & 0.150          \\
                                   & F-score & -              & 0.445          & 0.315          & 0.238          & 0.170          & 0.255          & 0.416          & 0.307          \\ \hline
    \multirow{3}{*}{NICE-SLAM\cite{zhuNICESLAMNeuralImplicita}}     
                                   & Acc.↓   & 0.303          & 0.119          & 0.116          & 0.216          & \underline{0.103}    & 0.156          & 0.464          & 0.211          \\
                                   & Comp.↓  & 0.456          & 0.042          & 0.070          & 0.199          & \textbf{0.089} & 0.163          & 0.045          & 0.152          \\
                                   & F-score & 0.221          & 0.554          & 0.411          & 0.402          & \underline{0.400}    & 0.273          & 0.401          & 0.380          \\ \hline
    \multirow{3}{*}{Ours}          
                                   & Acc.↓   & \textbf{0.081} & \textbf{0.068} & \textbf{0.067} & \textbf{0.079} & \textbf{0.078} & \underline{0.113}    & \textbf{0.121} & \textbf{0.087} \\
                                   & Comp.↓  & \underline{0.071}    & \underline{0.041}    & \underline{0.045}    & \underline{0.062}    & \underline{0.122}    & \underline{0.085} & \textbf{0.033} & \underline{0.066} \\
                                   & F-score & \textbf{0.559} & \textbf{0.661} & \textbf{0.646} & \textbf{0.643} & \textbf{0.496} & \textbf{0.557} & \textbf{0.656} & \textbf{0.604} \\ 
    \bottomrule
    \end{tabular}
  
\caption{\textbf{Quantitative Comparison on Reconstruction.} We perform the mapping evaluation on 7 indoor sequences and report results of three metrics including accuracy (Acc.), completion (Comp.) and F-score. The failure cases are marked as ``-''.}
\label{table:mesh}
\end{table*}

\begin{table*}[!t]
    \centering
    \small
    \begin{tabular}{lccccccc}
    \toprule
    Scene Name                                                    & Kitchen        & Sofa           & Office         & Reception      & Living room    & Office2        & Sofa2          \\ \hline
    ORB-SLAM3~\cite{campos2021orb3}                               & \textbf{0.054} & -              & \textbf{0.017} & \textbf{0.025} & -              & \textbf{0.022} & -              \\
    ORB-SLAM3(with $\widetilde D$)~\cite{campos2021orb3}          & 0.082          & \textbf{0.035}          & 0.019          & 0.049          & -              & 0.058          & -              \\
    \hline
    KinectFusion~\cite{izadi2011kinectfusion}                     & -              & 0.146          & 0.209          & 0.157          & -          & 0.321          & 0.125          \\
    ElasticFusion~\cite{whelanElasticFusionRealtimeDense2016}     & 0.253          & 0.110          & 0.070          & 0.193          & 0.530          & 0.121          & 0.146          \\
    BundleFusion~\cite{daiBundleFusionRealtimeGlobally2017}       & 0.176          & 0.102          & 0.135          & \underline{0.101}    & -              & 0.163          & \underline{0.120}    \\
    iMAP~\cite{sucar2021imap}                                     & -              & 1.658          & 0.338          & 0.648          & 0.679          & 0.344          & 0.214          \\
    NICE-SLAM~\cite{zhuNICESLAMNeuralImplicita}                   & 0.745          & 0.144          & 0.155          & 0.251          & \underline{0.289}    & 0.228          & 0.421          \\
    Ours                                                          & \underline{0.113}    & \underline{0.081}    & \underline{0.056}    & 0.114          & \textbf{0.200} & \underline{0.101}    & \textbf{0.085} \\
    \bottomrule
    \end{tabular}
    
    \caption{\textbf{Camera Tracking Results.} ATE RMSE [m] (↓) is used as the evaluation metric. The failure cases are marked as ``-''. For the variations of ORB-SLAM3, we only mark the best one in each sequence. Our approach outperforms all the other methods except for ORB-SLAM3~\cite{campos2021orb3}. However, ORB-SLAM3 fails on 3 of the 7 sequences due to the textureless indoor environment while our approach tracks successfully on all of the sequences.
    } 
    \label{table:traj}
\end{table*}

\section{Experiments}
\label{sec:experiment}

\subsection{Experimental Setup}
\noindent
\textbf{Datasets.} Since no public SLAM benchmarks take the L5 signals as input, we build a data capture device as in~\cite{li2022deltar} and use it to create an indoor SLAM dataset, which contains 7 sequences captured in 5 typical indoor scenes.
We record color images (640$\times$480), L5 signals and depth maps for each sequence.
Note that we only use the color images and L5 signals as the input to our SLAM system. The depth maps are used to obtain the ground truth 6-DoF camera poses and 3D surface mesh. We follow the automated capture pipeline proposed in ScanNet~\cite{dai2017scannet} to compute the ground truth data.

\noindent\textbf{Baselines.}
We compare our method with two categories of baselines:  (a) learning-based SLAM methods including iMAP~\cite{sucar2021imap} (re-implemented by~\cite{zhuNICESLAMNeuralImplicita}) and NICE-SLAM~\cite{zhuNICESLAMNeuralImplicita}; (b) traditional SLAM methods including ORB-SLAM3~\cite{campos2021orb3}, KinectFusion~\cite{izadi2011kinectfusion}, ElasticFusion~\cite{whelanElasticFusionRealtimeDense2016} and BundleFusion~\cite{daiBundleFusionRealtimeGlobally2017}. 
For the methods that only support RGB-D input, including iMAP, NICE-SLAM, KinectFusion, ElasticFusion and BundleFusion, we use the RGB image and the predicted depth map by~\cite{li2022deltar} as the system input. 
For ORB-SLAM3, we evaluate it in both the RGB version (given only RGB image as input) and the RGB-D version (using the predicted depth, marked as $\widetilde D$).
Note that none of these methods can work well with only raw L5 zone-level depths or take it as additional inputs.
We evaluate both the scene reconstruction and the camera tracking results. Since ORB-SLAM3 cannot output dense models, we exclude it from the mapping evaluation.

\begin{table*}[!ht]
\centering
\small
\begin{tabular}{cccccccccc}
\toprule
E    & M    & Kitchen    & Sofa       & Office     & Reception  & Living room                     & Office2    & Sofa2      & Avg.       \\ \hline
     &      & 0.203      & 0.088      & 0.063      & 0.143      & 0.216      & 0.120      & 0.094      & 0.132      \\
\checkmark &      & 0.135      & 0.085      & 0.060      & 0.144      & 0.211      & 0.108      & 0.089      & 0.119      \\
\checkmark &  \checkmark &  \textbf{0.113} &  \textbf{0.081} &  \textbf{0.056} &  \textbf{0.114} &  \textbf{0.200} &  \textbf{0.101} &  \textbf{0.085} &  \textbf{0.107} \\
\bottomrule
\end{tabular}
\caption{\textbf{Ablation Study.} We explore the efficiency of multi-modal scene representation (marked as ``M'') and L5 signal temporal filtering  (marked as ``E''). We use ATE RMSE [m] (↓) as the evaluation metric.} 
\label{table:ablation}
\end{table*}

\begin{table}[!t]
\centering
\small
\begin{tabular}{cccc}
\toprule
\multicolumn{1}{l}{}                                                              &               & Tracking & Mapping \\ \hline
\multirow{3}{*}{\begin{tabular}[c]{@{}c@{}}Learning-based \\ Methods\end{tabular}} & iMAP~\cite{sucar2021imap}          & 101 ms   & 448 ms  \\
                                                                                  & NICE-SLAM~\cite{zhuNICESLAMNeuralImplicita}     & 470 ms   & 1300 ms \\
                                                                                  & Ours          & 116 ms   & 380 ms  \\ \hline
\multirow{2}{*}{\begin{tabular}[c]{@{}c@{}}Classical \\ Methods\end{tabular}}      & ORB-SLAM3~\cite{campos2021orb3}     & 31 ms    & 159 ms  \\
                                                                                  & ElasticFusion~\cite{whelanElasticFusionRealtimeDense2016} & 31 ms    & -       \\ 
\bottomrule
\end{tabular}
\caption{\textbf{Runtime Comparison.} \new{Different from NICE-SLAM~\cite{zhuNICESLAMNeuralImplicita} that measures one iteration of optimization, we report the averaging total runtime for tracking and mapping respectively for better comparison with classic methods.} ElasticFusion does not have an explicit mapping process thus it is labeled as ``-''.}
\label{table:runtime}
\end{table}

\noindent\textbf{Implementation Details.}
Our SLAM system is executed on a desktop PC equipped with an Intel i7-9700K CPU and an NVIDIA RTX 3090 GPU.
In all our experiments, we set the grid sizes to [3, 6, 24, 96] cm
and the number of sampling pixels $M$ to 5000.
Our method is implemented using PyTorch~\cite{paszke2019pytorch} with ADAM~\cite{kingma2014adam} optimizer. 
The decoders, multi-modal feature grids, and camera poses are trained with learning rates of 0.001, 0.01, and 0.0005, respectively.
The loss weights are set to 10 for $\lambda_{c}$ and $\lambda_{sdf}$ and 1 for the others.
Inspired by instant-ngp~\cite{instant_ngp}, we use the tiny-cuda-nn~\cite{tiny-cuda-nn} library to implement the proposed multi-modal grid encoding, which significantly accelerates the optimization process.
\new{We follow DELTAR~\cite{li2022deltar} to pre-train the multi-modal depth prediction network on NYU-V2 with simulated L5 signals.}
For more details, please refer to the supplementary material.

\subsection{Evaluation of Mapping and Tracking}
\noindent\textbf{Mapping.}
We use three metrics to evaluate the reconstruction result including accuracy (Acc.), completion (Comp.) and F-score following the previous work~\cite{sunNeuralReconRealTimeCoherent2021a}. For detailed descriptions of these metrics, please refer to the supplementary material.
As shown in Table~\ref{table:mesh}, our method outperforms all the baseline methods by a large margin (ours 0.604 vs. the second-best 0.476 in terms of F-score). We also show qualitative results in Fig.~\ref{fig:mesh}. Our method is able to produce high-quality 3D models with smooth surfaces and high accuracy. It is easy to notice that our reconstruction result has much fewer artifacts and noisy points. \new{Since NICE-SLAM~\cite{zhuNICESLAMNeuralImplicita} relies on high-quality depth input, its performance is poor given the noisy and unreliable depth input.}

\noindent\textbf{Tracking.}
We use ATE RMSE~\cite{sturm2012benchmark} to evaluate the camera tracking. As shown in Table~\ref{table:traj}, our method outperforms 
all other methods
except for ORB-SLAM3~\cite{campos2021orb3}.
However, ORB-SLAM3 is much less robust and tracks lost on 3 of the 7 sequences, while our method successfully tracks all the sequences.
This is because that ORB-SLAM3, as a keypoint-based method, struggles in texture-less regions, which are commonly seen in indoor environments.
Given the predicted depth, ORB-SLAM3 can reduce the lost ratio from 42.8\% to 28.5\% but leads to worse camera tracking results since the predicted depth is not accurate enough. 

\noindent\textbf{Runtime Analysis.}
\new{
We compare the runtime of tracking and mapping in Table~\ref{table:runtime} for both classical methods and learning-based methods. 
The runtime of our method is similar to iMAP, but it is faster than NICE-SLAM due to our implementation optimization inspired by instant-ngp~\cite{instant_ngp}.
}

\subsection{Ablation Study}
\label{sec:ablation}
In this section, we first study the effectiveness of using pixel-wise depth prediction as additional supervision,
then investigate the importance of the proposed multi-modal feature grid representation and the temporal filtering technique.

\begin{figure}[!t]
\begin{center}
\includegraphics[width=1.0\linewidth]{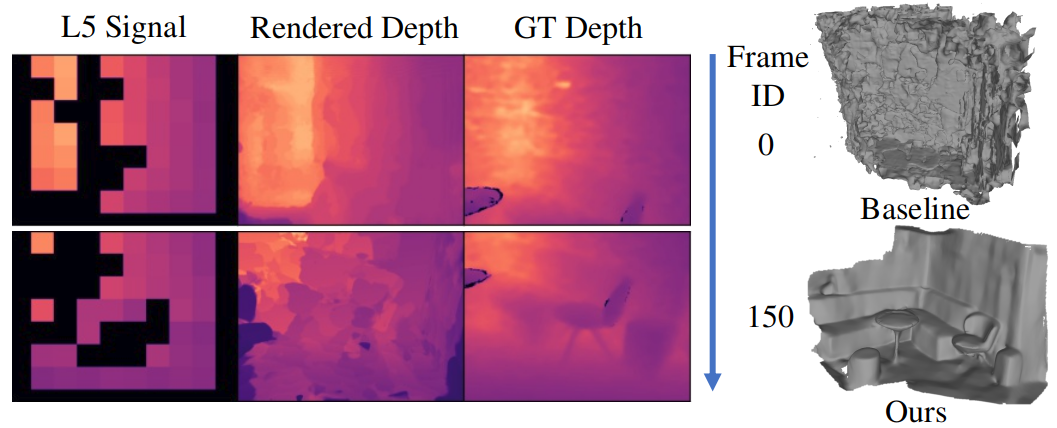}
\end{center}
\caption{\textbf{Results without Pixel-Wise Depth Supervision.} It cannot guarantee plausible reconstruction results without the pixel-wise depth maps as additional supervision.}
\label{fig:raw_signal}
\end{figure}

\noindent\textbf{Impact of the Pixel-Wise Depth Supervision.}
We try using only the L5 raw signals and RGB images for supervision.
\new{
Removing the pixel-wise depth supervision makes the optimization problem harder, which is more obvious when the problem size grows. 
In Fig.~\ref{fig:raw_signal} we show the rendered depth when optimizing the first frame and the first 150 frames. The latter contains serious errors.  
As a result, it leads to a distorted reconstruction, \ie, the ``Baseline'' result in the upper right.
}

\noindent\textbf{Impact of the Multi-Modal Feature Grid.}
We verify the effectiveness of our multi-modal feature grid representation (denoted as ``M'')
in terms of camera tracking and show the result in Table~\ref{table:ablation}.
With the multi-modal feature grid representation, we are able to use raw L5 signal as supervision. As a result, we can optimize the camera poses on both pixel-level and zone-level, leading to better trajectory accuracy.

\begin{figure}[!t]
\begin{center}
\includegraphics[width=1.0\linewidth]{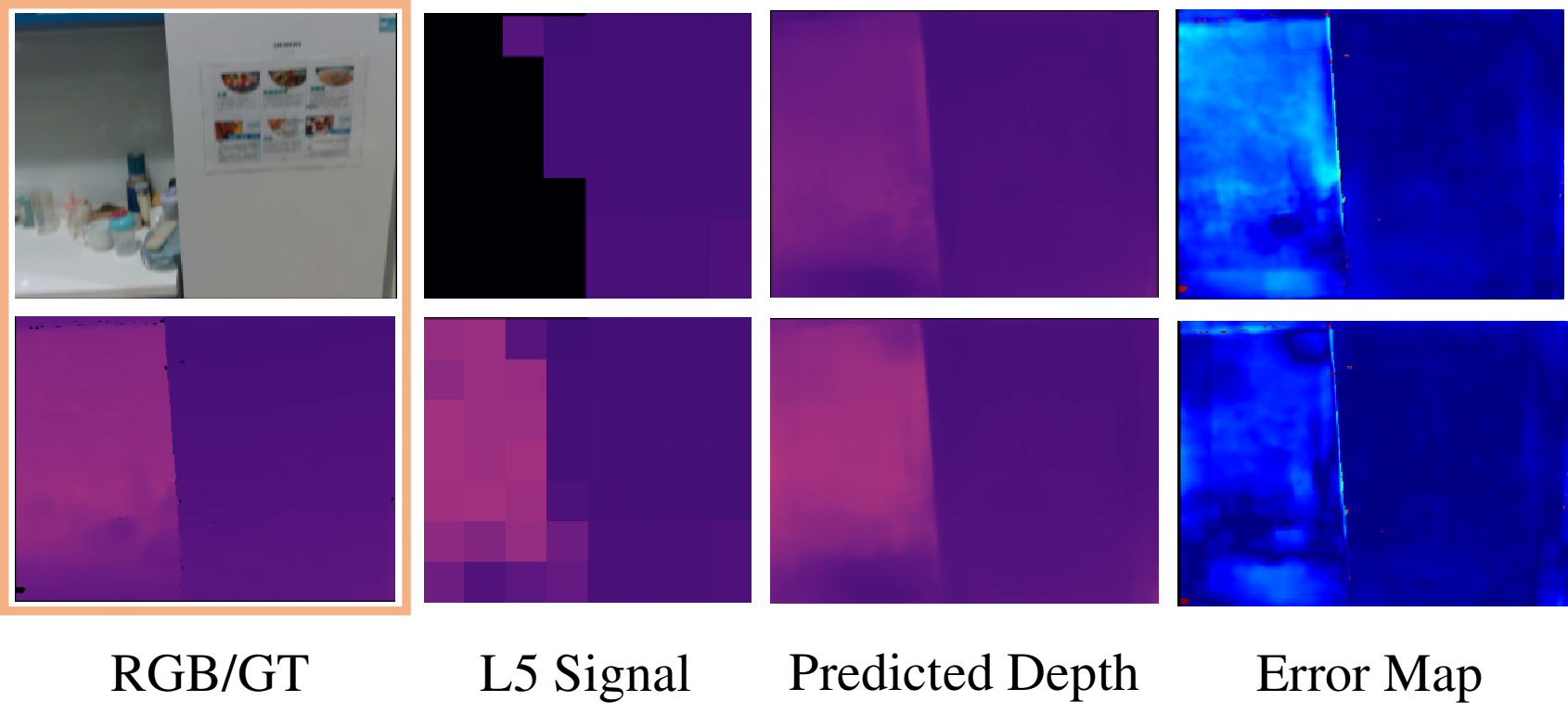}
\end{center}
\caption{\textbf{Qualitative Study of the Effect of Temporal Filtering on Depth Prediction.} 
\new{We compare the depth prediction result obtained through two methods: solely employing raw L5 signals (top row) and incorporating our temporal filtering technique (bottom row).}
Our temporal filtering technique enhances the raw L5 signals and fills up the missing regions, leading to a better depth prediction result.}
\label{fig:depth_refine}
\end{figure}

\noindent\textbf{Impact of the Temporal Filtering.} 
In Fig.~\ref{fig:depth_refine} we show the qualitative comparison with and without the proposed temporal filtering.
It can be seen that the large portion of missing L5 signals leads to severe errors on the corresponding regions of the predicted depth map. The error reduces significantly with the proposed temporal filtering technique.

\begin{table}[!t]
\small
\begin{tabular}{lllllll}
\toprule
    &    & {$\delta_1\uparrow$}  & {$\delta_2\uparrow$}& {$\delta_3\uparrow$}   & {REL$\downarrow$}  & {RMSE$\downarrow$}           \\ \hline
\multicolumn{1}{c}{\multirow{2}{*}{\begin{tabular}[c]{@{}c@{}} H \end{tabular}}} & w/o E          & 0.773          & 0.910          & 0.969          & 0.159          & 0.513          \\
\multicolumn{1}{c}{}                                                                               & w E & \textbf{0.925} & \textbf{0.972} & \textbf{0.983} & \textbf{0.093} & \textbf{0.366} \\ \hline

\multirow{2}{*}{\begin{tabular}[c]{@{}l@{}} N \end{tabular}}                     & w/o E            & 0.954          & 0.989          & 0.997          & 0.065          & 0.151          \\
       & w E & \textbf{0.956} & \textbf{0.992} & \textbf{0.998} & \textbf{0.065} & \textbf{0.145} \\ \bottomrule
       
\end{tabular}
\caption{\textbf{Ablation Study on Temporal Filtering for Depth Prediction.} ``H''/``N'' stands for the hard/normal cases and ``E'' represents the temporal filtering. We report the quantitative result of depth maps predicted from L5 raw signals and our refined signals.} 
\label{table:depth_refine}
\end{table}

\begin{figure}[!t]
\begin{center}
\includegraphics[width=.95\linewidth]{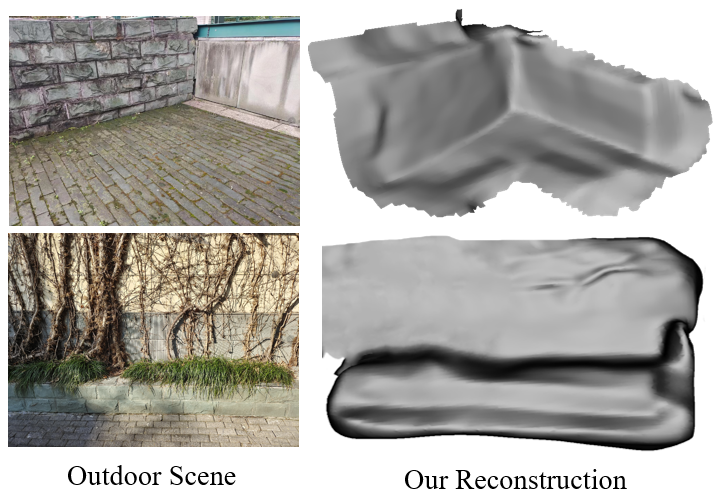}
\end{center}
\caption{\textbf{Qualitative Results of Outdoor Reconstruction.} The proposed method can recover a complete scene model in outdoor scenes but missing more details than that in indoor due to the poor L5 signals.}
\label{fig:outdoor}
\end{figure}

We also study the quantitative impact of temporal filtering (marked as ``E'') on both the depth prediction (Table~\ref{table:depth_refine}) and the camera tracking (Table~\ref{table:ablation}).
We divide the testing sequences into normal and hard cases according to the quality of L5 signals.
As shown in Table~\ref{table:depth_refine}, the temporal filtering technique improves the predicted depth in general and significantly in hard cases. Such an improvement benefits the final SLAM system and leads to more accurate camera tracking, as shown in Table~\ref{table:ablation} (marked as ``E'').
Please refer to the supplementary material for the definition of the normal/hard cases and the depth evaluation metrics.

\subsection{Performance in Outdoor Scenes}

Currently, we focus on indoor scenes as other NeRF-based SLAM systems~\cite{sucar2021imap,zhuNICESLAMNeuralImplicita} due to the optical interference limitation of depth sensors. Actually, the sunlight has a stronger interference on the L5 sensor for its low-power emitter. In Fig.~\ref{fig:outdoor}, we show two examples of outdoor scene reconstruction using our proposed methods. The recovered mesh is complete but missing more details than that in indoor scenes due to the poor L5 signals.

\section{Conclusion}
We introduce a novel dense visual SLAM framework working with RGB cameras and light-weight ToF sensors using neural implicit scene representation. 
To accommodate this new input modality, we propose a novel multi-modal feature grid that enables both zone-level rendering for the ToF sensors and pixel-level rendering for other high-resolution signals. To guarantee robust tracking and mapping,  we exploit a per-pixel depth prediction as additional supervision, which is further improved by a novel temporal filtering strategy.
Our experiments demonstrate that the proposed method can provide accurate camera tracking and high-quality reconstruction result on indoor scenes. 
Similar to other NeRF-based RGB-D SLAM systems, as future work we plan to further improve the system to overcome the  limitation of ToF sensors in outdoor scenarios and make it efficient enough to run on mobile robots. 

\noindent \textbf{Acknowledgements:}
\new{This work was partially supported by NSF of China (No. 62102356).}

{\small
\bibliographystyle{ieee_fullname}
\bibliography{main}
}

\clearpage

\appendix

\renewcommand\thesection{\Alph{section}}
\renewcommand\thetable{\Alph{table}}
\renewcommand\thefigure{\Alph{figure}}

{\noindent \Large \bf Supplementary Material\par}

\vspace{0.5em}

\renewcommand\thesection{\Alph{section}}
\renewcommand\thetable{\Alph{table}}
\renewcommand\thefigure{\Alph{figure}}

In this supplementary document, we provide more implementation details in Sec.~\ref{sec:imple-details},
describe more details about the light-weight ToF sensor we used in Sec.~\ref{sec:L5-principle}, more details on the evaluation in Sec.~\ref{sec:metrics}, and show more qualitative results in Sec.~\ref{sec:results}. 
Finally, we discuss our limitations in Sec.~\ref{sec:limitation}.
More qualitative results can be found in our supplementary video.

\section{More Implementation Details}
\label{sec:imple-details}
\subsection{Multi-Modal Implicit Scene Representation}
As described in Sec.~\textcolor{red}{3.2} of our main paper, we propose a new multi-model implicit scene representation with multi-level grid features. In the multi-model feature grids, the dimension of features for the geometry grid at each level is 4. Since we only encode color at the finest level, color features are encoded with a dimension of 6. 
We use the dense grid from tiny-cuda-nn~\cite{tiny-cuda-nn} library to implement the multi-level feature grids for acceleration. For both geometry and color features, we use small MLPs with two hidden layers consisting of 32 neurons as decoders.  

The input dimension of the geometry decoder is 16, corresponding to the total geometry feature size. When decoding pixel-level depth, all the input neurons are active; while decoding zone-level depth, only 8 of the 16 neurons corresponding to the zone-level features are active. The input dimension of the RGB decoder is 9, including 6 for the color features extracted from the feature grid and 3 for the view direction vector.

\subsection{Loss Function}
In this section, we provide more details of the SDF supervision and the SDF regularization terms used in the mapping process.

\noindent\textbf{SDF Supervision Term.}
Apart from supervising the rendered depth, we also supervise the intermediate SDF prediction.
Following~\cite{ortizISDFRealTimeNeural2022, wangGOSurfNeuralFeature2022, azinovicNeuralRGBDSurface}, we approximate the ground-truth SDF supervision $ b({\bf{x}}) = \widetilde D[u,v] - d $ based on our depth predictions. It is noticeable that this approximation is an upper bound of the ground-truth SDF value. For near-surface points, the differences between the ground-truth SDF value and the approximation are expected to be smaller. As a result, we apply the following near-surface loss for the near-surface points (${|\widetilde D[u, v] - d| <  = t}$) as in \cite{ortizISDFRealTimeNeural2022}:
\begin{equation}
{\ell _{sdf}}({\bf{x}}) = |\phi_{pix} ({\bf{x}}) - b({\bf{x}})|,
\end{equation}
where $\phi_{pix}$ represents the pixel-level SDF value of point $\bf{x}$ as mentioned in Sec.~\textcolor{red}{3.2} of our main paper.
The truncation threshold $t$ is a hyper-parameter and we set it to 16cm.
For points far from the surface (${|\widetilde D[u, v] - d| > t}$), we apply the following loss to encourage free space prediction as \cite{ortizISDFRealTimeNeural2022}:
\begin{equation}
{\ell _{fs}}({\bf{x}}) = \max \left( {0,{e^{ - \beta \phi_{pix} ({\bf{x}})}} - 1,\phi_{pix} ({\bf{x}}) - b({\bf{x}})} \right),
\end{equation}
where $\beta$ is a hyper-parameter controlling the exponential penalty term when the SDF prediction is negative, and we set $\beta = 5$ in our experiment.

\noindent\textbf{SDF Regularization Term.}
To alleviate the ill-posed problem
in under-constrained regions, we employ two additional regularization terms on the SDF prediction:
Eikonal regularization $ \ell _{eik} $ and smoothness regularization $ \ell _{smooth} $. 

Eikonal regularization is widely used in previous works~\cite{wangNeuSLearningNeural2021, ortizISDFRealTimeNeural2022}
, encouraging the prediction to approximate a signed distance function.
This regularization can help propagate the SDF field from the near-surface regions to free space. For a given point $ \bf{x} $, the Eikonal regularization is applied via the loss $ \ell _{eik} $:
\begin{equation}
{\ell _{eik}}({\bf{x}}) = {(1 - ||\nabla \phi_{pix} ({\bf{x}})||)^2}.
\end{equation}

Since we do not have a high-quality depth map as input like~\cite{zhuNICESLAMNeuralImplicita, sucar2021imap}, we also add a smoothness regularization that encourages nearby points to have similar normal direction as in~\cite{wangGOSurfNeuralFeature2022}:
\begin{equation}
{\ell _{smooth}}({\bf{x}}) = ||\nabla \phi_{pix} ({\bf{x}}) - \nabla \phi_{pix} ({\bf{x}} + \epsilon)|{|^2},
\end{equation}
where $ \epsilon $ is a small perturbation with a random direction and length of $\delta_s$. We set $ \delta_s $ to 4mm empirically.

\noindent\textbf{Experimental Settings.}
We use the first 60 frames to do initialization. During the initialization process, the frames are sequentially added to the optimization process every $ N_a = 100$ iterations, and after all frames are added, an extra $ N_e = 300 $ iterations of optimization are performed. We set the number of tracking iterations $N_t$ to 50, the number of mapping iterations $N_m$ to 150, the number of sampling pixels $ M $ to 5000, the number of sampling zones $Z$ to 500, the number of neighbor frames $N_n$ to 60, the window size $ W_s $ to 30 and implement a hierarchical sampling strategy similar to NeuS~\cite{wangNeuSLearningNeural2021} to obtain the sampling points along the ray. We first sample $N_c = 96$ coarse samples and add 12 samples at each step based on weights computed from the previously sampled points.

\section{Light-Weight ToF Sensor Details}
\label{sec:L5-principle}
\begin{figure}[!t]
\begin{center}
    \includegraphics[width=0.75\linewidth]{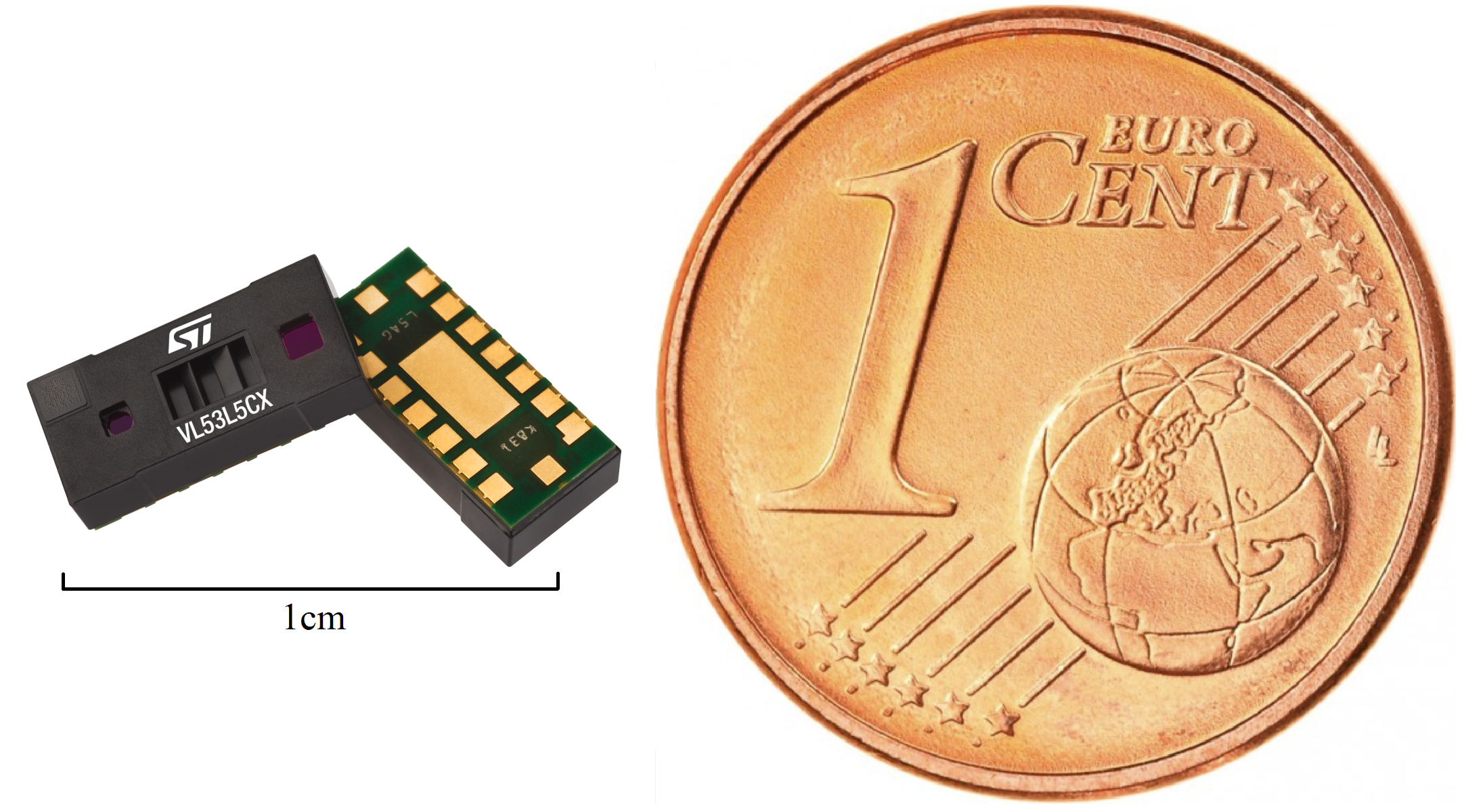}
\end{center}
\caption{\textbf{VL53L5CX ToF Sensor.} The sensor is extremely small in size (6.4 $\times$ 3.0  $\times$ 1.5 mm) and runs at a fairly low power consumption (about 200mW). We compare the size of the VL53L5CX sensor with a one cent euro coin in the figure.}
\label{fig:l5}
\end{figure}

\begin{table}[!t]
\centering
\scalebox{0.85}{
\begin{tabular}{ccc}
\toprule
\multicolumn{1}{l}{} & \multicolumn{1}{l}{VL53L5CX(ours)} & \multicolumn{1}{l}{Apple LiDAR} \\ \hline
Cost                 & \$2-3                              & $\sim$\$20                      \\
Resolution           & 8×8                                & 256×192                         \\
Power                & 0.2W                               & 3-4W                            \\
Main Usage           & Autofocus                          & AR \& VR                        \\ \bottomrule
\end{tabular}
}
\caption{\textbf{Comparison with Apple LiDAR.} }
\label{table:sensors}
\end{table}
In this paper, we use ST VL53L5CX~\cite{l5_page} (shown in Fig.~\ref{fig:l5}, denoted as L5) as a typical representative of light-weight ToF sensors. Comparing with common commodity level depth sensors (e.g. Intel RealSense, apple LiDAR, etc.), light-weight ToF sensors are a magnitude lower in power consumption and price. We compare the basic information between L5 and apple LiDAR in Table~\ref{table:sensors}.
L5 outputs depth in the resolution of $8\times 8$, and each measures the depth distribution in a large zone. Its diagonal field-of-view (FoV) is 63$^{\circ}$, similar to common monocular cameras.
Unlike the conventional ToF sensor 
that provides per-pixel high-resolution (usually higher than 0.03 megapixels) depth map,
L5 provides depth distribution measurements within zones in an extremely low resolution.

L5 emits infrared rays and measures depth based on the time taken for the wave to bounce back to the emitter. However, due to the light-weight electronic design, L5 is only able to give statistical information about the depth in a large zone. The depth distribution is initially obtained by counting the number of photons returned in each discretized range of time and then fitted with a Gaussian distribution model to compress the raw information due to its tight bandwidth limit. 

For each zone, L5 also returns a status code to show whether the measurement in that zone is valid. If the number of photons received in a zone is too small or the measurements are unstable, the corresponding zone will be marked as invalid. More details can be found on STMicroelectronics's webpage\footnote{https://www.st.com/content/st\_com/en/premium-content/premium-content-time-of-flight.html}.

\section{Evaluation Details}
\label{sec:metrics}

\subsection{Mesh Culling}
Implicit methods can usually complete the scene geometry for unseen regions. For a fair comparison between implicit and explicit methods, we cull surfaces that are not observed inside any camera frustums or occluded by other objects.                                                                                                                                                                                      

\subsection{Depth Metrics}
\begin{figure*}[!t]
\begin{center}
    \includegraphics[width=0.85\linewidth]{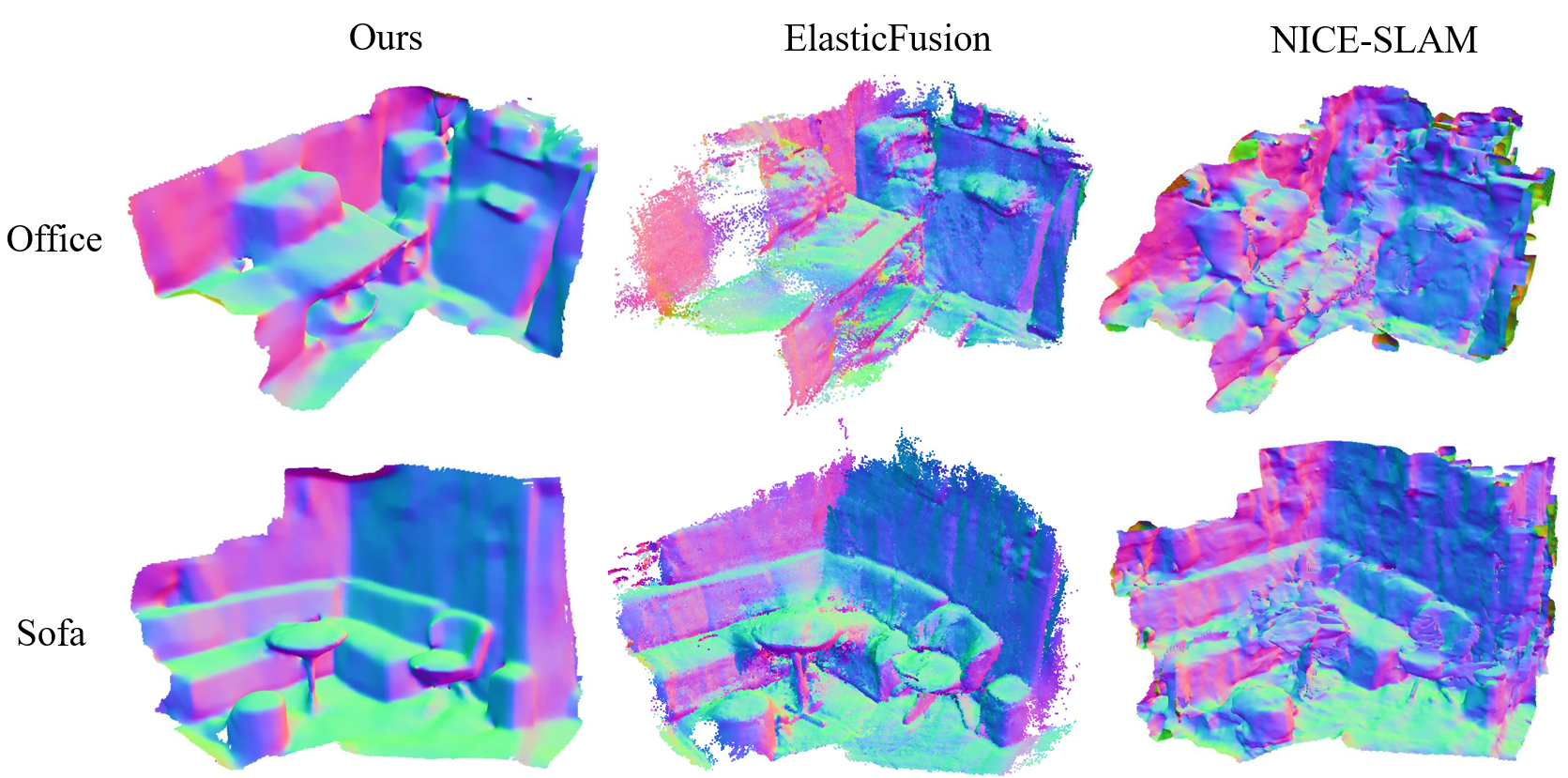}
\end{center}
\caption{\textbf{More Reconstruction Results.} We show the final mesh generated by NICE-SLAM~\cite{zhuNICESLAMNeuralImplicita}, ElasticFusion~\cite{whelanElasticFusionRealtimeDense2016} and our method. The mesh is visualized with the vertex normal. Compared to other methods, our method recover cleaner and sharper scene geometry.}
\label{fig:normal}
\end{figure*}

\subsection{Reconstruction Metrics}
We evaluate the quality of the scene reconstruction using the following standard metrics where $p$ and $p^*$ are the vertices in generated mesh $P$ and GT mesh $P^*$ respectively: 
\begin{itemize}
  \item Accuracy (Acc.):
  \begin{equation}
    \frac{1}{{|P|}}\sum\limits_{p \in P} {{{\min }_{p* \in {P^*}}}\left\| {p - {p^*}} \right\|}.
  \end{equation}
  \item Completeness (Comp.):
  \begin{equation}
    \frac{1}{{|{P^*}|}}\sum\limits_{p* \in {P^*}} {{{\min }_{p \in P}}\left\| {p - {p^*}} \right\|}.
  \end{equation}
  \item Precision (Prec.):
  \begin{equation}
    \frac{1}{{|P|}}\sum\limits_{p \in P} {{{\min }_{p* \in {P^*}}}\left\| {p - {p^*}} \right\| < 0.05}. 
  \end{equation}
   \item Recall (Recal.):
  \begin{equation}
    \frac{1}{{|{P^*}|}}\sum\limits_{p* \in {P^*}} {{{\min }_{p \in P}}\left\| {p - {p^*}} \right\| < 0.05}.
  \end{equation}
  \item F-score:
  \begin{equation}
      \frac{{2 \times {\rm{ Prec}}{\rm{. }} \times {\rm{ Recal}}{\rm{.}}}}{{{\rm{ Prec}}{\rm{.  +  Recal}}{\rm{.}}}}.
  \end{equation}
\end{itemize}
In general, F-score is considered as the most proper metric to evaluate the quality of the scene reconstruction \cite{sunNeuralReconRealTimeCoherent2021a} since both the accuracy and completeness of the reconstruction are considered.

We evaluate the performance of the depth prediction using the following standard metrics where $\hat{d_i}$ represents predicted depth, $d_i$ represents ground truth depth, and N is the number of valid ground truth values:
\begin{itemize}
  \item Threshold Accuracy ($\delta_i$ with i=1,2,3):
  \begin{equation}
      \frac{\sum_{i=1}^N[max(\frac{\hat{d_i}}{d_i},\frac{d_i}{\hat{d_i}}) < 1.25^i]}{N},
  \end{equation}
  where $[]$ denotes Iverson brackets.
  \item Mean Absolute Relative Error (REL):
  \begin{equation}
      \frac{1}{N}\sum_{i=1}^{N}\frac{|\hat{d_i}-d_i|}{d_i}.
  \end{equation}
  \item Root Mean Square Error (RMSE):
  \begin{equation}
      \sqrt{\frac{1}{N}\sum_{i=1}^{N}(\hat{d_i}-d_i)^2}.
  \end{equation}
\end{itemize}

In the ablation study about the temporal filtering technique for depth prediction, we do the evaluation separately for normal and hard cases. Here, we give a detailed definition of the division criteria. For the normal case, the raw L5 signal is of normal quality, leading to relatively accurate depth prediction. While for the hard case, %
the raw L5 signal is noisy or has large amounts of missing data. 
We classify a prediction as a normal case if its RMSE error is less than 0.4; otherwise, we regard it as a hard case.

\begin{figure}[!t]
\begin{center}
    \includegraphics[width=0.67\linewidth]{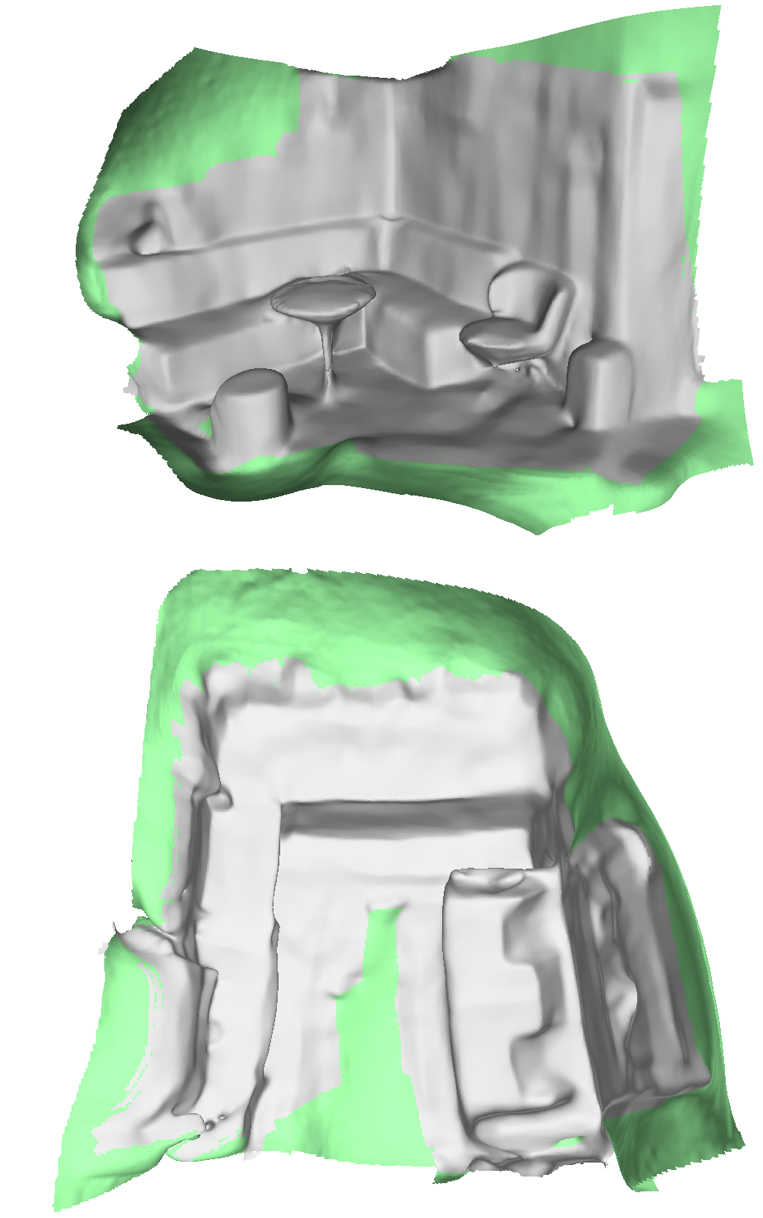}
\end{center}
\caption{\textbf{Geometry Forecast.} The white-colored area is the region with observations, while the green-colored area represents the unseen but forecasted region. It is noticeable that our method can generate reasonable mesh even in the unseen region.}
\label{fig:forecast}
\end{figure}

\begin{figure}[!t]
\begin{center}
    \includegraphics[width=0.8\linewidth]{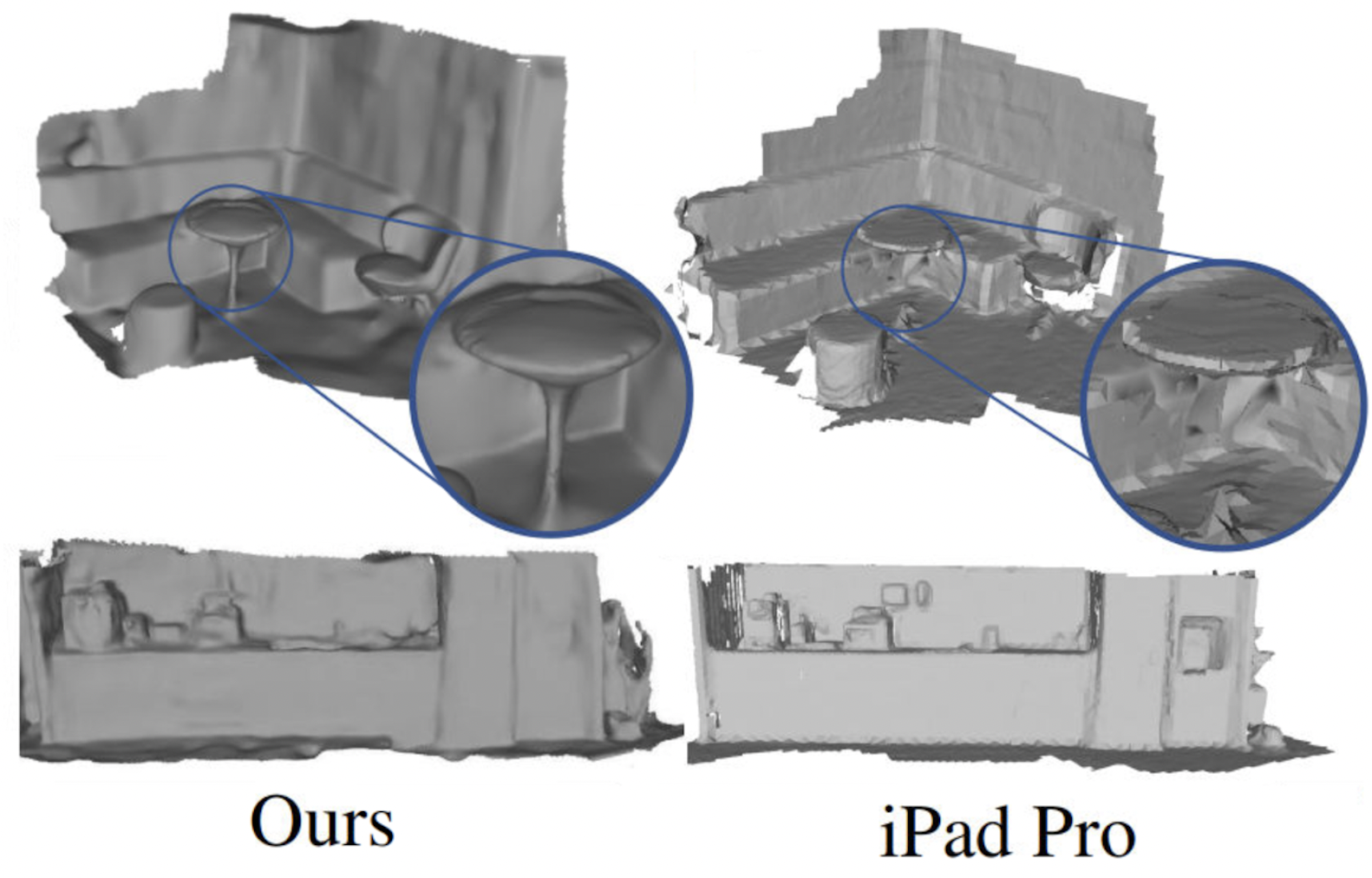}
\end{center}
\caption{\textbf{Comparison to a Smartphone Scan.}}
\label{fig:ipadpro}
\end{figure}

\section{More Qualitative Results}
\label{sec:results}

\noindent\textbf{Mesh Visualization with Vertex Normal.}
To better highlight the differences in reconstruction quality, we provide additional visualizations in Fig.~\ref{fig:normal} using the vertex normal to color the mesh. It is noticeable that our approach outperforms the others and produces high-quality scene reconstruction results. 

\vspace{1.0em}

\noindent\textbf{Geometry Forecast.}
Our method is able to make reasonable predictions in unseen regions thanks to the multi-modal implicit scene representation. As shown in Fig.~\ref{fig:forecast}, the hole in the floor is well filled, and the walls are correctly expanded to unobserved regions.

\vspace{1.0em}

\noindent\textbf{Comparison to a Smartphone Scan.}
We used an iPad Pro (with ``3D Scanner App'') to re-scan two testing scenes under as closely identical conditions as possible to ours.
As shown in Fig.~\ref{fig:ipadpro}, our method can achieve reconstruction results comparable to the iPad Pro using a much cheaper light-weight ToF and outperforms iPad Pro on thin objects thanks to the multi-modal implicit scene representation.

\section{Limitations}
\label{sec:limitation}
Firstly, the range of a light-weight ToF sensor is usually limited to several meters and the sunlight has a strong interference on the sensor, so our method currently focuses on indoor scenes. We plan to further improve the system to overcome the limitation of ToF sensors in outdoor scenarios.
Secondly, the computational overhead of the proposed method is still relatively high for mobile platforms. In future work, we plan to further reduce the computational burden and make it efficient enough to run on mobile robots.
At last, we also plan to add semantic information into our system %
for high-level scene understanding.

\end{document}